\pdfoutput=1

\documentclass[11pt]{article}

\usepackage[preprint]{acl}

\usepackage{times}
\usepackage{latexsym}

\usepackage[T1]{fontenc}

\usepackage[utf8]{inputenc}

\usepackage{microtype}

\usepackage{inconsolata}

\usepackage{graphicx}

\usepackage{hyperref}
\usepackage{url}
\usepackage{graphicx}
\usepackage{booktabs}
\usepackage{multirow}
\usepackage{wrapfig}
\usepackage{xcolor}
\usepackage{amsmath}
\usepackage{xfrac}
\usepackage{titlesec}
\usepackage{tcolorbox}
\usepackage{ifthen}
\usepackage{xspace}
\usepackage{etoolbox}
\usepackage{listings}
\usepackage{tikz}
\usepackage{enumitem}
\usepackage{svg}

\newcommand{\method}{{WirelessMathBench}}
\newcommand{\NumOfQ}{{587}}

\newcommand{\examplebox}[2]{
    \begin{tcolorbox}[colframe=darkgray, colback=white, boxrule=0.5pt, width=1.0\linewidth,fontupper=\small,fontlower=\small\textit,title=#1.]
        #2
    \end{tcolorbox}\vspace{-1.2em}
}

\title{\textbf{\method}: A Mathematical Modeling Benchmark for LLMs in Wireless Communications}

\newcommand{\EEE}{\textsuperscript{1}} 
\newcommand{\KU}{\textsuperscript{2}} 

\author{
{Xin Li\EEE $\quad$ Mengbing Liu\EEE  $\quad$ Li Wei\EEE  $\quad$  Jiancheng An\EEE} \\
\textbf{Mérouane Debbah\KU $\quad$  Chau Yuen\EEE}\\
{\small \EEE School of Electrical and Electronic Engineering (EEE), Nanyang Technological University, Singapore}\\
{\small \KU Department of Computer and
Information Engineering, Khalifa University, Abu Dhabi, UAE}\\
{\tt\small \href{https://lixin.ai/WirelessMathBench}{https://lixin.ai/WirelessMathBench}}
}

\begin{document}

\maketitle

\begin{abstract}
Large Language Models (LLMs) have achieved impressive results across a broad array of tasks, yet their capacity for complex, domain-specific mathematical reasoning---particularly in wireless communications---remains underexplored. 
In this work, we introduce \method\, a novel benchmark specifically designed to evaluate LLMs on mathematical modeling challenges to wireless communications engineering. 
Our benchmark consists of 587 meticulously curated questions sourced from 40 state-of-the-art research papers, encompassing a diverse spectrum of tasks ranging from basic multiple-choice questions to complex equation completion tasks, including both partial and full completions, all of which rigorously adhere to physical and dimensional constraints.
Through extensive experimentation with leading LLMs, we observe that while many models excel in basic recall tasks, their performance degrades significantly when reconstructing partially or fully obscured equations, exposing fundamental limitations in current LLMs. Even DeepSeek-R1, the best performer on our benchmark, achieves an average accuracy of only 38.05\%, with a mere 7.83\% success rate in full equation completion.
By publicly releasing \method\ along with the evaluation toolkit, we aim to advance the development of more robust, domain-aware LLMs for wireless system analysis and broader engineering applications.
\end{abstract}

\section{Introduction}
\label{sec:intro}

\begin{figure}[!ht]
    \centering
    \includegraphics[width=\linewidth]{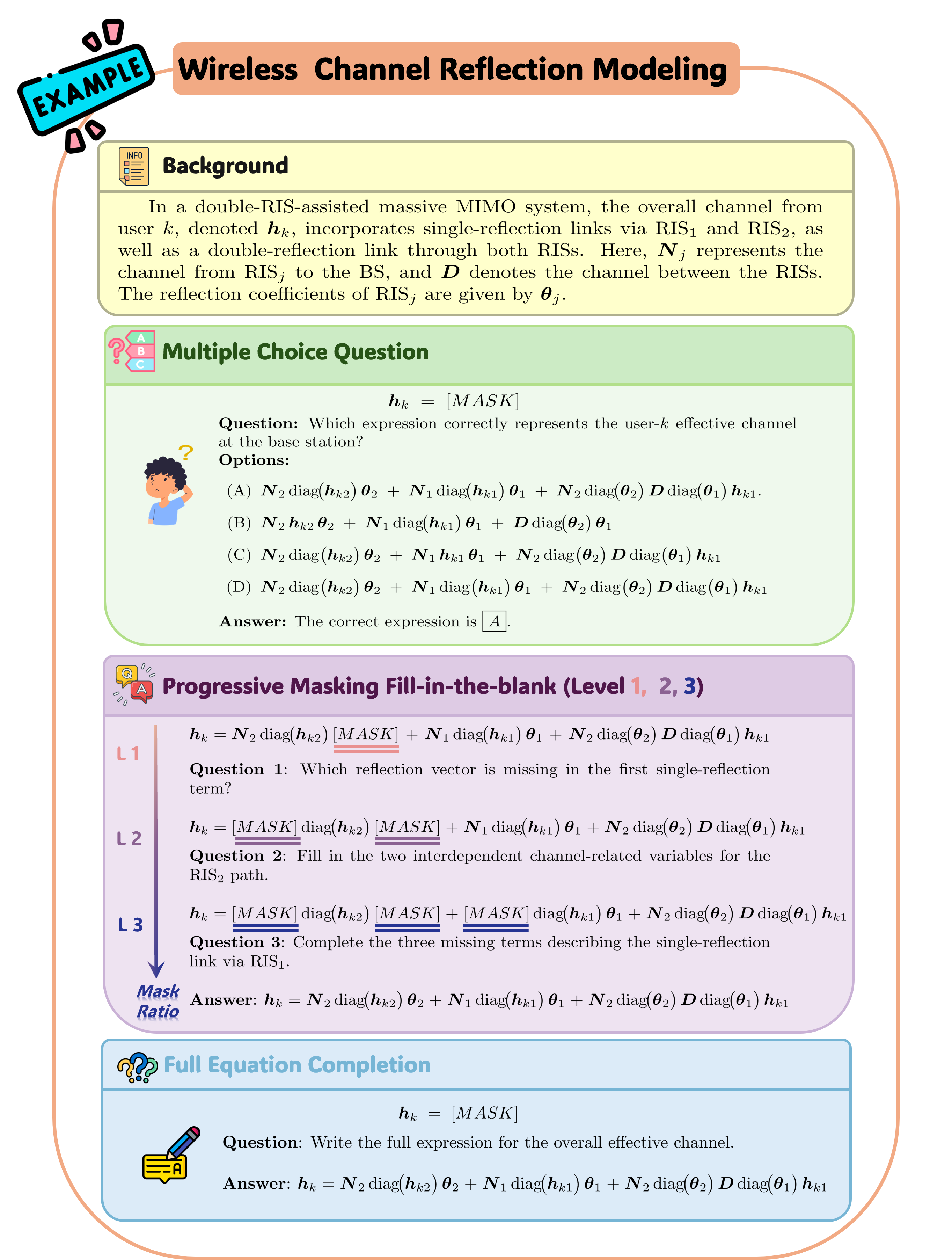}
    \caption{Example task from \method\: a system model derivation from wireless communications literature. The derivation progresses from a multiple-choice question to progressive mask completion questions, and finally to the full formula derivation, testing the model's ability to reason through complex channel reflections and matrix operations.}
    \label{fig:teaserfigure}
\end{figure}

Large Language Models (LLMs) have recently demonstrated groundbreaking performance across a diverse range of natural language tasks---from general language understanding~\citep{brown2020language, wang2018glue, wang2019superglue} and code generation~\citep{lu2021codexglue} to elementary mathematical reasoning~\citep{cobbe2021gsm8k, hendrycks2020measuring}. 
Advanced models such as OpenAI-o1~\citep{openai2024learning} and DeepSeek-R1~\citep{guo2025deepseek} have further extended these capabilities, especially when supplemented with chain-of-thought strategies that enable clear, step-by-step solution processes. 
Nevertheless, despite these notable achievements, current state-of-the-art LLMs still encounter significant difficulties when tackling highly intricate problem statements. In particular, tasks that demand deep conceptual insights, rigorous validation of physical feasibility, and the careful management of tightly interrelated parameter sets continue to pose formidable challenges~\citep{mirzadeh2024gsm,zhang2023ai,he2024olympiadbench}.

In many engineering fields---wireless communications in particular---mathematical modeling is indispensable. The design and analysis of modern wireless systems require not only accurate numerical computation but also precise symbolic derivations that honor strict physical and dimensional constraints.
Tasks such as channel estimation~\citep{JSAC_2013_Yin_A, liu2022deep, CL_2023_An_A}, beamforming~\citep{TVT_2023_Chu_Joint, TSP_2004_Spencer_Zero,liu2025beamforming}, and multi-antenna system~\citep{WC_2020_Huang_Holographic, WC_2024_An_Stacked,WC_2024_Zheng_Mobile,liu2025air} involve intricate matrix operations, multi-stage derivations, and domain-specific lexicon. Even minor errors in symbolic manipulation can lead to significant performance degradation or non-compliance with industry standards~\citep{TCOM_2013_Bjornson_A}.

Although recent work has leveraged LLMs for technical definition retrieval in wireless communications~\citep{shao2024wirelessllm, maatouk2023teleqna, zou2024telecomgpt, maatouk2024tele}, few studies have directly addressed the challenges associated with multi-step derivations and symbolic manipulation in this specialized domain. This observation raises a broader question: \emph{To what extent are LLMs capable of emulating the mathematical derivations and analytical typically by an engineer or researcher in the field of real wireless communications?}

To bridge this gap, we introduce \method\, a comprehensive benchmark specifically designed to test LLMs on the real-work wireless engineering orientation mathematical reasoning.
\method\ comprises \NumOfQ~high-quality questions sourced from 40 state-of-the-art papers, each carefully annotated and validated by domain experts to ensure accuracy.
These questions span a variety of system models (e.g., Multiple-Input and Multiple-Output (MIMO), Non-orthogonal multiple access (NOMA), Reconfigurable Intelligent Surfaces (RIS) ) and problem settings (e.g., channel estimation, beamforming), encompassing multiple-choice, fill-in-the-blank, and open-ended questions at various levels.
Table\ref{tab:benchmark_comparison} highlights key differences between \method\ and other math benchmarks, ours is the only dataset of expert difficulty level and contains real-world engineering problems.
Figure~\ref{fig:teaserfigure} illustrates how a single math formula escalates from a basic multiple-choice query to a fully masked equation derivation, reflecting the complexity of real-world wireless system analysis.

Our extensive experiments show that while leading LLMs perform well on simpler tasks (e.g., multiple-choice questions with over 75\% accuracy), their performance drops dramatically on advanced derivation tasks (progressive masking and complete equations). Even the strongest model we evaluated, DeepSeek-R1~\citep{guo2025deepseek}, only manages a 7.83\% success rate in fully masked derivations, underscoring a fundamental gap between current LLM capabilities and the complex demands of wireless systems analysis.

By publicly releasing \method\ along with its evaluation toolkit, we aim to spur progress toward LLMs that are not only fluent in natural language but also capable of rigorous, domain-specific mathematical reasoning. 
We envision that \method\ will serve as a catalyst for innovation in mathematical reasoning capabilities, domain-adaptive pre-training techniques, and advanced thought-chaining strategies, ultimately propelling LLMs toward more robust scientific and engineering problem-solving.

\section{Related Work}
\label{sec:related}

\begin{table*}[!t] 
\centering 
\resizebox{.9\linewidth}{!}{
\begin{tabular}{lcccc} 
    \toprule 
    \textbf{Benchmark} & \textbf{Diffuculty Level} &\textbf{QuestionType}& \textbf{Real Engineering Tasks}&\textbf{\#Test Size}\\
    \midrule 
    GSM8K~\citep{cobbe2021gsm8k}& Elementary School &OE& No  &1,319\\
 MATH~\citep{hendrycks2021measuring}& High
School &OE& No  &5,000\\
 OCWCourses~\citep{lewkowycz2022solving}& University &OE& No  &272\\
 MMMU~\citep{yue2024mmmu}& University &MC,OE& No  &1983\\
 OlympiadBench~\citep{he2024olympiadbench}& Competition &OE& No  &8,476\\ 
    
    \textbf{\method} & \textbf{Expert} &\textbf{MC, FB, OE}& \textbf{Yes}&\textbf{587}\\ 
    \bottomrule 
\end{tabular} 
}
\caption{
Comparison of representative mathematical benchmarks with \textbf{\method}. 
Existing datasets largely focus on elementary, high school, or Olympiad-level problems in purely theoretical contexts, while \textbf{\method} targets real-world, expert-level engineering tasks under strict dimensional and physical constraints. 
We note that open-ended (OE) tasks typically require free-form answers, MC indicates multiple-choice, and FB refers to fill-in-the-blank.
}
\label{tab:benchmark_comparison} 
\end{table*}

\paragraph{General-Purpose LLM Benchmarks.}  
In recent years, rapid advancements in LLMs---exemplified by models such as 
GPT-3~\citep{brown2020language}, GPT-4~\citep{openai2023gpt4}, LLaMA~\citep{touvron2023llama}, Gemini~\citep{team2023gemini}, and DeepSeek-R1~\citep{guo2025deepseek}---have spurred extensive evaluations on benchmarks like 
GLUE~\citep{wang2018glue}, SuperGLUE~\citep{wang2019superglue}, and GSM8K~\citep{cobbe2021gsm8k}.
However, despite covering a broad spectrum of linguistic tasks, they typically lack the depth and specificity required to evaluate rigorous mathematical modeling or the domain-specific symbolic reasoning needed for complex technical applications.

\paragraph{Mathematical Reasoning Benchmarks.}  
A parallel research stream has focused on the mathematical and symbolic reasoning abilities of LLMs. 
Early mathematical benchmarks~\cite{amini2019mathqa, cobbe2021gsm8k, koncel2016mawps, ling2017program,hendrycks2021measuring} evaluate models on elementary arithmetic, algebra, and calculus problems.
Recently, as the complexity of the problem increases, some benchmarks introduce competition-level problems that combine mathematical logic and background knowledge~\citep{yu2023metamath, hendrycks2020measuring, arora2023have,frieder2024mathematical}.
For more advanced mathematical reasoning, datasets like MMMU~\citep{yue2024mmmu}, OCWCourses~\citep{lewkowycz2022solving} and U-MATH~\citep{chernyshev2024u} focuses on university-level mathematics problems.
MiniF2F~\citep{zheng2022miniff}, AlphaGeometry~\citep{trinh2024solving}, OlympiadBench~\citep{he2024olympiadbench}, and MathOdyssey~\citep{fang2024mathodyssey} go further to Olympiad-level problems that require more advanced mathematical reasoning.
Yet, these datasets do not capture the unique constraints or specialized notations found in applied domains like wireless communications.

\paragraph{Domain-Specific Benchmarks.}  
To overcome the limitations of general-purpose evaluations, several domain-specific benchmarks have been developed for tasks that demand technical precision and specialized reasoning.
For example, customized benchmarks have been developed for 
legal document analysis~\citep{guha2024legalbench}, 
chemical property inference~\citep{guo2023can}, 
and scientific reasoning~\citep{lu2022learn,wang2024scibench, sun2024scieval}. 
To evaluate LLMs in more specialized domain tasks, recent works have introduced benchmarks like MLAgentBenchmark~\citep{huang2024mlagentbench}, which evaluates LLMs' ability to solve machine learning tasks, AI-Researcher~\citep{si2024llmideas} evaluate can LLMs generate research ideas, and SWE-Bench~\citep{jimenez2023swe} evaluate LLMs' ability to solve real-world software engineering tasks.
These studies highlight that an in-depth evaluation of LLMs in specialized fields reveals that LLMs have strong potential in different professional fields.

\paragraph{LLMs vs. Symbolic Solvers in Engineering.}
Symbolic mathematics tools (e.g., Mathematica~\cite{wolfram2003mathematica}, SymPy~\cite{meurer2017sympy}) have traditionally dominated computation-intensive engineering workflows. While these tools excel at manipulating well-defined symbolic expressions, they fundamentally lack the capability to translate unstructured natural language descriptions into formal mathematical representations~\cite{androutsopoulos1995natural,manning1999foundations}. This limitation is particularly pronounced in wireless communications, where system models require integration of domain knowledge, physical constraints, and specialized notation. LLMs potentially address this gap through their ability to process natural language specifications and generate corresponding mathematical formulations~\cite{openai2023gpt4,cobbe2021gsm8k}---a capability critical for real-world engineering applications but not comprehensively evaluated by existing benchmarks.

\paragraph{LLMs in Wireless Communications.}  
Wireless communications impose stringent requirements on mathematical precision, particularly for tasks such as channel estimation, interference management, and beamforming~\citep{TIT_2008_Cadambe_Interference, TSP_2011_Shi_An, JSAC_2010_Gesbert_Multicell}. 
Some preliminary works have explored the use of LLMs in wireless contexts, focusing on domain-specific knowledge extraction and basic recall of technical standards~\citep{maatouk2023teleqna,shao2024wirelessllm,maatouk2024tele}.
Notably, TelecomGPT~\citep{zou2024telecomgpt} has extended LLM capabilities to higher-level tasks like wireless-specific code generation and formula completion. 
However, these early works usually emphasize knowledge retrieval or summarization, without considering testing what tasks LLMs can accomplish in actual wireless communication engineering systems.

In this work, we introduce \method\ to address these gaps.
Unlike existing wireless or purely mathematical benchmarks, \method\ offers tasks that systematically combine multiple-choice questions with progressively masked formula derivations, all drawn from state-of-the-art research papers.
The goal is to evaluate both symbolic reasoning and domain knowledge under realistic conditions, capturing the nuanced interplay of mathematical derivations and physical feasibility inherent in wireless communications.
By providing a diverse set of tasks and domain-informed evaluation metrics, \method\ aims to facilitate collaborative advances in both LLMs and wireless communication engineering, ultimately enabling more powerful AI-assisted solutions for next-generation wireless networks.

\begin{figure*}[!htbp]
    \centering
    \includegraphics[width=.8\linewidth]{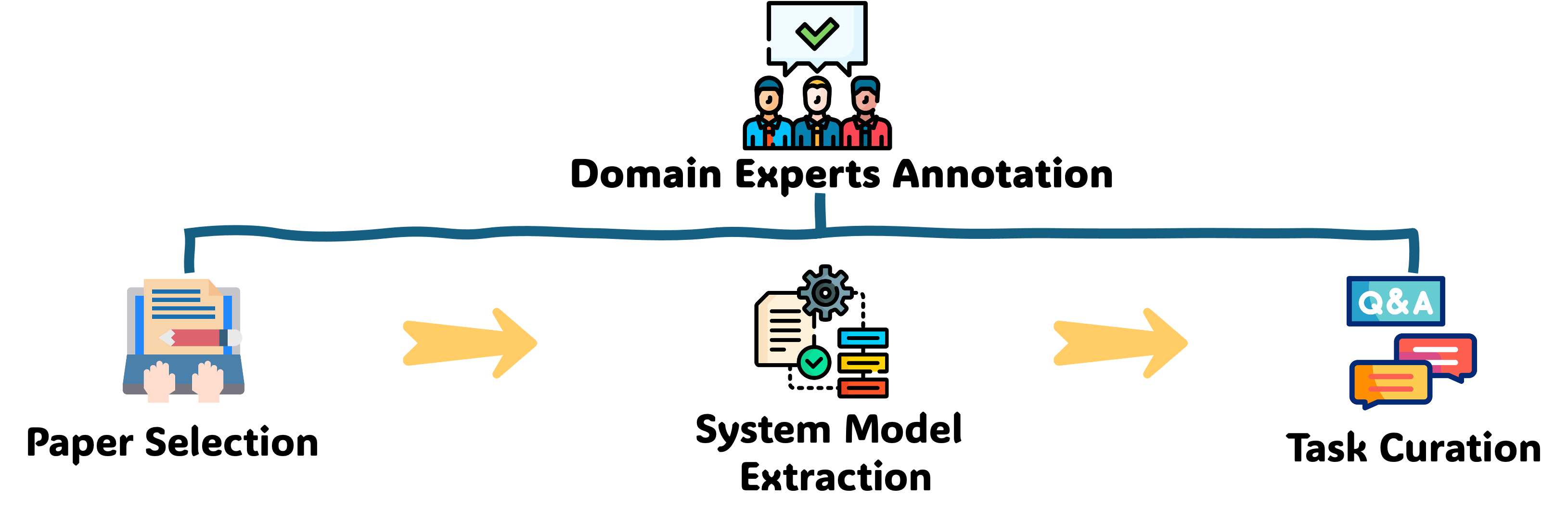}
    \caption{Overview of the data collection and annotation pipeline for \method\. The process involves selecting high-quality research papers, extracting system models from papers, curating tasks of varying complexity levels, and reviewing each task for clarity and correctness.}
    \label{fig:data_collection}
\end{figure*}

\section{The \method\ Benchmark}
\label{sec:method}

\begin{table}[tbp]
    \centering
    \resizebox{0.9\linewidth}{!}{
    \begin{tabular}{lcr} 
        \toprule
        \textbf{Category Type} & \textbf{Topic Category} & \textbf{Number of Papers} \\ 
        \midrule
        \multirow{7}{*}{Model-based}& RIS & 19\\
 & MIMO&12\\
 & UAV &6\\
 & ISAC &6\\
        & Satellite& 4\\ 
        & SIM & 3\\ 
        & NOMA& 2\\ 
        \midrule
        \multirow{6}{*}{Problem-based}& Beamforming& 18\\
 & Channel Estimation &12\\
 & Performance Analysis&8\\
 & Trajectory Design&5\\
 & Power Allocation&5\\
 & Resource Management&4\\ 
        \midrule
        \multicolumn{2}{l}{\textbf{Total}} & \textbf{40}\\
        \bottomrule
    \end{tabular}
    }
    \caption{
        Distribution of the \method\ benchmark papers according to model-based and problem-based categories, along with their respective topic areas. A total of 40 papers are included, covering key themes in wireless communications. Note that some papers may span multiple topic categories.
    }
    \label{tab:topic_distribution}
\end{table}

In this section, we present \method, a new benchmark specifically designed to evaluate LLMs on mathematical modeling tasks within wireless communications. 
We begin by discussing the rationale behind our benchmark design (Section~\ref{subsec:design_principles}), followed by the details of our data collection and annotation pipeline (Section~\ref{subsec:data_collection}). 
We then explain how we construct questions of varying complexity levels, as well as our progressive masking methodology (Section~\ref{subsec:task_design}).

\subsection{Design Principles}
\label{subsec:design_principles}

The creation of \method\ is motivated by two core observations.
First, recent work shows that LLMs can effectively assist humans in highly specialized tasks~\citep{guha2024legalbench, guo2023can, lu2022learn}, underscoring their potential when provided with sufficient domain context.
Second, LLMs have demonstrated the capacity to handle increasingly difficult mathematics, including Olympiad-level challenges~\citep{he2024olympiadbench, fang2024mathodyssey}.
These findings suggest a substantial opportunity to push the limits of LLMs in areas where complex, domain-specific mathematics---such as wireless communications---plays a central role.

Building on these insights, \method\ is designed around two key principles:

\begin{enumerate}[leftmargin=*]
\item \textbf{Real-World Complexity.}
Each task is sourced directly from peer-reviewed research, reflecting the authentic modeling challenges faced in wireless systems.
\item \textbf{Multi-Tiered Progression.}
Tasks range from basic multiple-choice questions to fully masked derivations, providing graduated levels of difficulty that capture both foundational knowledge and advanced reasoning.
\end{enumerate}

\subsection{Data Collection and Annotation}
\label{subsec:data_collection}

As illustrated in Figure~\ref{fig:data_collection}, the data collection and annotation process for \method\ involves four main steps: paper selection, system model extraction, task curation, and domain expert review.

\begin{figure}[t]
    \centering
    \includegraphics[width=\linewidth]{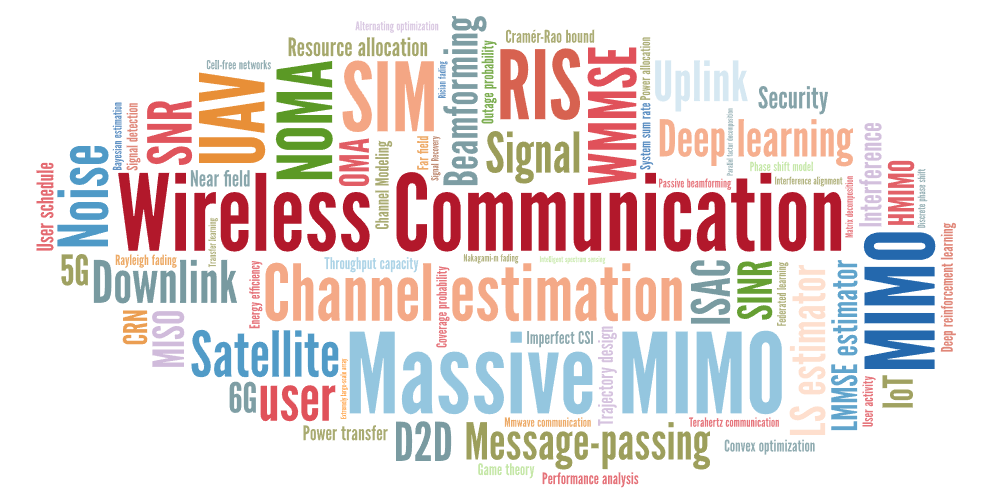}
    \caption{A word cloud illustrating the most frequent keywords in the \method\ benchmark, which reflects the range of wireless communication topics covered.}
    \label{fig:question_distribution}
\end{figure}

\paragraph{Paper Selection and Coverage.} To capture the authentic complexities of wireless communications, we begin by identifying high-impact papers from top-tier publication venues that are freely accessible on arXiv.
Table~\ref{tab:topic_distribution} summarizes the coverage of model-based and problem-based categories; in total, we select 40 papers spanning core techniques (e.g., MIMO, NOMA, RIS) and research focuses (e.g., channel estimation, beamforming).
We emphasize works that feature nontrivial mathematical derivations---such as optimization formulations and multi-stage channel modeling---over those limited to empirical or simulation-based heuristics.
A summary of the high-frequency keywords across the dataset is shown in Figure~\ref{fig:question_distribution}, highlighting the diverse wireless communication topics covered in the benchmark.
We aim to include tasks that reflect the symbolic depth and physical constraints that are indispensable for real-world wireless engineering and state-of-the-art wireless research.

\paragraph{System Model Extraction.}
Our pipeline starts by applying a specially designed LLM template that systematically scans each research paper, isolating key mathematical expressions and relevant contextual descriptions. 
This initial extraction step is semi-automated: 
LLMs produce a structured draft containing a concise overview of the system model, assumptions, and principal formulas. 
Subsequently, domain experts review and refine these drafts, ensuring that the extracted material is both accurate (i.e., symbolically consistent with the original text) and self-contained (i.e., providing enough background to be understood independently). 
To mitigate potential data contamination concerns, experts deliberately reformulate paper contexts in original language, restructure equation presentations, and avoid word-for-word reproductions of problem statements. These transformations ensure that our task formulations differ substantially from source materials that might appear in LLM training corpora, thereby requiring genuine mathematical reasoning rather than mere reproduction of memorized content.
This hybrid method combines the scalability of automated extraction with meticulous expert checks, ensuring the resulting text is accurate, symbolically consistent, and sufficiently self-contained for subsequent tasks while maintaining necessary differentiation from publicly available sources.

\paragraph{Task Curation.}
Once the system models have been extracted, we systematically transform them into precise question-answer pairs via a three-step process:

\begin{enumerate}[leftmargin=*]
    \item \textbf{Identify Core Equations:} We select representative formulas from each paper---such as channel gain expressions, optimization objectives, or multi-hop path-loss derivations---that encapsulate critical wireless engineering challenges.
    \item \textbf{Construct Questions:} For each core equation, we generate questions at varying levels of difficulty.  This includes:  (a) multiple-choice questions targeting definitions or partial operations;  (b) progressively masked fill-in-the-blank questions that require incremental reasoning; and  (c) full equation completion tasks that demand derivation of the entire expression.
    \item \textbf{Annotate and Review:} Each question is accompanied by contextual notes, and domain experts validate correctness and clarity. When necessary, we refine notation or provide short explanations to ensure the questions are self-contained and can be tackled without external references 
\end{enumerate}

\paragraph{Domain Expert Review.}
Lastly, all questions undergo a multi-round review by senior wireless researchers.
They verify notation accuracy and domain applicability (e.g., check for appropriate dimensionality, and coherent modeling assumptions), and remove any ambiguous or misleading content.
The remaining problems constitute the final \method\ dataset: a set of carefully selected tasks that embody typical mathematical derivations in advanced wireless communication engineering and research.
A detailed description of our expert validation methodology, including reviewer qualifications and the verification workflow, is provided in Appendix~\ref{app:expert_validation}.

\subsection{Task Design and Masking Strategies} \label{subsec:task_design}

To full evaluate the capabilities of LLMs at different levels of difficulty in mathematical modeling of wireless communications, \method\ incorporates three distinct task types. 
Each question leverages real-world system equations derived from state-of-the-art research papers, ensuring that the benchmark reflects both conceptual diversity and practical engineering relevance.
At the same time, each independent question is accompanied by a brief description of the relevant wireless scenario (e.g., UAV relay or multi-antenna beamforming), providing the necessary domain and scenario background information.

\paragraph{Multiple-Choice Questions (MCQs).}
These questions require the solver to select the correct mathematical expression from a set of closely related distractors, with each MCQ carefully designed to test the model's ability to recognize and recall key elements of wireless system modeling. For example, a typical MCQ may present several equations for a wireless channel, of which only one formula satisfies both the correct dimensions and the physical constraints of the system under consideration.

\paragraph{Progressively masked fill-in-the-blank questions. } 
In this task, a system model formula is progressively presented in a partially masked form across three different masking levels. 
Each progressively masked instance is treated as an independent subproblem, requiring the model to infer and reconstruct the missing information at each stage. 
The masking levels range from isolated single-variable omissions to multi-variable occlusions, with varying degrees of accompanying prompt text to provide contextual guidance.

\paragraph{Full Equation Completion (FEC) question.}
For the most challenging question, the full equation is entirely hidden.
The solver is provided with only a succinct description of the wireless scenario (for example, a base station-relay-user link with specific path loss characteristics) and must derive the complete expression from first principles. This task assesses the model's ability to reconstruct the entire derivation---from fundamental definitions (like channel gain or fading coefficients) to the final expression---while ensuring dimensional accuracy and adherence to domain-specific constraints (such as path loss exponent and transmit power limits). 
It represents the level of performance expected from a human expert in wireless communications.

In summary, by combining MCQs, progressively masked tasks, and full equation completions, \method\ offers a comprehensive, fine-grained evaluation of a model's capability to perform both symbolic reasoning and domain-specific derivations in wireless communications.

\section{Experiments}
\label{sec:experiments}

We now present a comprehensive evaluation of \method, focusing on how leading LLMs handle wireless communications--specific mathematical modeling. We begin by detailing our experimental setup (Section~\ref{subsec:exp-setup}), including prompt design and model baselines, then discuss our main results (Section~\ref{subsec:main-results}), and conclude with an error analysis that highlights key challenges (Section~\ref{subsec:error-analysis}).

\begin{table}[!tb]
    \centering
    \resizebox{\linewidth}{!}{
    \begin{tabular}{lcc}
    \toprule
    \textbf{Model}& \textbf{Source}& \textbf{Size}\\
    \midrule

 OpenAI-o1&~\citep{openai2024learning}& unknown\\
 DeepSeek-R1&~\citep{guo2025deepseek} & 671B \\
 GPT-Family&~\citep{openai2024gpt4o,openai2023gpt4}&unknown\\
  DeepSeek-V3&~\citep{deepseekai2024deepseekv3technicalreport} & 671B\\
    Gemini-Famliy&~\citep{google2024gemini2} & unknown \\
 Qwen2.5-Math&~\citep{yang2024qwen25mathtechnicalreportmathematical} &7B, 72B\\
    LLaMA        &~\citep{grattafiori2024llama3herdmodels} & 8B, 70B\\
    LLaMA-3-8B-Tele   &~\citep{maatouk2024tele} & 8B\\
 Mistral-7B&~\citep{jiang2023mistral}&7B\\
    \bottomrule
    \end{tabular}
    }
    \caption{LLMs evaluated on \method.}
    \label{tab:llms}
\end{table}

\begin{table*}[!tb]
    \centering
    \begin{tabular}{lcccccc}
    \toprule
    \multirow{2}{*}{\textbf{Model}} & \multirow{2}{*}{\textbf{MCQ}}& \multicolumn{3}{c}{\textbf{Progressive Masking Filling}}&\multirow{2}{*}{\textbf{FEC}}&\multirow{2}{*}{\textbf{Avg. Acc}}\\
    
 \cmidrule(lr){3-5}&   & Level 1& Level 2& Level 3& & \\
    \midrule
 DeepSeek-R1& 76.00\%& \textbf{60.00\%}& \textbf{34.91\%}& \textbf{12.50\%}& \textbf{7.83\%}& \textbf{38.05\%}\\
 OpenAI-o1& 66.40\%& 59.17\%& 32.17\%& 8.04\%& 6.96\%&34.55\%\\
 OpenAI-o1-mini& 66.40\%& 53.33\%& 29.57\%& 10.71\%& 4.35\%&32.87\%\\
    \midrule
 GPT-4o& 72.80\%& 42.50\%& 28.70\%& 6.25\%& 4.35\%&30.92\%\\
 GPT-4& 53.60\%& 38.33\%& 18.26\%& 3.57\%& 4.35\%&23.62\%\\
 GPT-3.5-turbo& 45.60\%& 7.50\%& 10.43\%& 1.79\%& 1.74\%&13.41\%\\
 DeepSeek-V3& \textbf{78.40\%}& 50.00\%& 24.35\%& 6.25\%& 6.96\%&33.19\%\\
    Gemini-2.0-flash& 71.20\%& 40.83\%& 24.35\%& 5.36\%& 4.35\%& 29.22\%\\
 Gemini-1.5-pro& 65.60\%& 43.33\%& 29.57\%& 9.82\%& 6.09\%&30.88\%\\
 Gemini-1.5-flash& 66.40\%& 37.50\%& 13.91\%& 2.68\%& 4.35\%&24.97\%\\
    \midrule
 Qwen2.5-Math-72B& 70.40\%& 37.50\%& 26.09\%& 7.14\%& 6.09\%&29.44\%\\
    LLaMA-3.3-70B        & 65.60\%& 38.33\%& 17.39\%& 2.68\%& 6.09\%& 26.02\%\\
    \midrule
 Qwen2.5-Math-7B& 58.40\%& 21.67\%& 6.96\%& 4.46\%& 1.74\%&18.82\%\\
    LLaMA-3-8B-Tele   & 40.80\%& 11.67\%& 4.35\%& 2.68\%& 0.87\%& 12.07\%\\
    LLaMA-3-8B        & 45.60\%& 10.83\%& 7.83\%& 2.68\%& 2.61\%& 13.91\%\\
 Mistral 7B& 38.40\%& 20.00\%& 4.35\%& 0.89\%& 0.87\%&12.90\%\\
    \bottomrule
    \end{tabular}
    \caption{Experimental results of state-of-the-art LLMs on \method. The table shows the performance of each model on MCQ, progressively masked filling and full equation completion tasks.}
    \label{tab:exp_main}
\end{table*}

\subsection{Experiment Setup}
\label{subsec:exp-setup}

\paragraph{Evaluation Workflow.}  
All experiments are conducted in a zero-shot setting using unified prompt templates across different question types for consistent evaluation. 
For each task in \method, we provide the corresponding prompt to each model and collect the answers it generates.
Our evaluation pipeline is now completed in two main ways. 
For multiple-choice questions, we directly extract the output answers and compare their consistency with the annotation results. 
For Progressive Masking Filling and Fully masked questions, since polynomials may have a certain number of possible answers, we use the help of LLMs (GPT-4o is selected in our experiment) to complete the evaluation, similar to\citep{fang2024mathodyssey, chernyshev2024u}. 
The overall performance is reported as the average accuracy of all tasks. 
Detailed prompt examples and scoring criteria are provided in the Appendix~\ref{app:prompt_examples}.

\paragraph{Baselines.}

Table~\ref{tab:llms} (in the main text) lists the principal models tested. We include leading reasoning models(e.g., DeepSeek-R1, OpenAI-o1), large-scale general-purpose LLMs (e.g., GPT-4, Gemini), and specialized models (e.g., Qwen2.5-Math) to capture a broad range of capabilities. For open-source models like LLaMA, we also explore domain-specific variants trained on a telecom corpus (e.g., LLaMA-3-8B-Tele) to gauge the benefit of targeted adaptation. 
All hyperparameters follow each model’s respective default or recommended settings, and no additional chain-of-thought prompting is provided beyond the standard instructions above.

\subsection{Main Results}
\label{subsec:main-results}

Table~\ref{tab:exp_main} presents the performance of sixteen LLMs across five metrics in \method: (1)~Multiple-choice Question (MCQ) accuracy, (2–4)~progressive masking fill-in at three difficulty levels (Level 1, Level 2, Level 3), (5)~Full  Equation Completions (FEC), and the overall average accuracy (Avg. Acc). 
Our key findings are summarized below:

\paragraph{Reasoning-Oriented Models Show Advantages.} 
Models that incorporate explicit chain-of-thought or advanced reasoning techniques---like DeepSeek-R1 and OpenAI-o1---consistently outperform simpler large-scale baselines. The average accuracy of \textbf{DeepSeek-R1} is 38.05\%, and the average accuracy of \textbf{OpenAI-o1} is 34.55\%, while the accuracy of other large-parameter models hovers around 30\%. This suggests that explicit reasoning strategies contribute substantially to managing multi-step symbolic derivations in wireless communications tasks. This performance gap stems primarily from reasoning models' ability to decompose complex mathematical operations into sequential sub-steps---a capability critical for the non-trivial matrix manipulations and dimensionality constraints inherent in wireless modeling tasks. As demonstrated in recent works~\citep{guo2025deepseek,shao2024deepseekmath}, models with explicit reasoning mechanisms excel at tasks requiring symbolic consistency across multiple operations, systematically tracking variables through transformations and validating intermediate results. 

\paragraph{Strong MCQ Performance but Rapid Decline in Derivations.}
Several models, including DeepSeek R1, V3, GPT-4, and Gemini-2.0, exceed 70\% accuracy on MCQs, showing that they can find the correct formula given background knowledge and given error options, indicating that they can understand the modeling process and matrix operations in the communication domain to some extent. 
However, these MCQ gains generally do not extend to more complex derivation tasks, where most models’ accuracy falls dramatically. For instance, \textbf{DeepSeek-V3} achieves the highest MCQ score at 78.40\%, drop to 6.25\% in Level 3 masking filling, and 6.96\% in FEC.

\paragraph{Progressive Masking Emphasizes Multi-Step Reasoning Gaps.}
When forced to reconstruct partially hidden expressions, model performance declines in proportion to the level of masking. When forced to reconstruct partially hidden expressions, model performance degrades with increasing levels of masking. Models with implicit reasoning logic significantly outperform the others, with \textbf{DeepSeek-R1} in particular leading on these tasks---achieving  60.00\% at Level 1 and 33.91\% at Level 2, suggesting more robust chaining of thoughts. However, even DeepSeek-R1 struggles at level 3 (12.50\%), highlighting the difficulty of maintaining symbolic coherence under heavily masked conditions.

\paragraph{Fully Masked Equation Completion Remains Challenging.}
Most models attain only single-digit accuracy (2–7\%) in the FEC task, where the entire equation is hidden. \textbf{DeepSeek-R1}’s 7.83\% and \textbf{OpenAI-o1}’s 6.96\% are the best in this category, but both remain low in absolute terms, indicating that fully reconstructing multi-step derivations without partial clues poses a significant challenge.

\paragraph{Domain-Focused Models Show Improvements.}
Models that are specifically tuned for mathematical reasoning---such as Qwen2.5-Math---demonstrate improved performance over other models with a similar parameter count, both in terms of overall average accuracy and on individual subtasks. In particular, \textbf{Qwen2.5-Math-72B} achieves an average accuracy of 29.44\%, which is on par with the average performance of most commercial models. 
However, fine-tuning general-purpose models like LLaMA to telecom-specific data (\emph{e.g.}, LLaMA-3-8B-Tele) yields only limited benefits. This is likely because the telecom fine-tuning data predominantly consists of wireless protocols, whereas the problems in \method\ require handling long contexts and performing high-level mathematical reasoning.

\subsection{Error Analysis}
\label{subsec:error-analysis}

\begin{figure}[t]
    \centering
    \includegraphics[width=\linewidth]{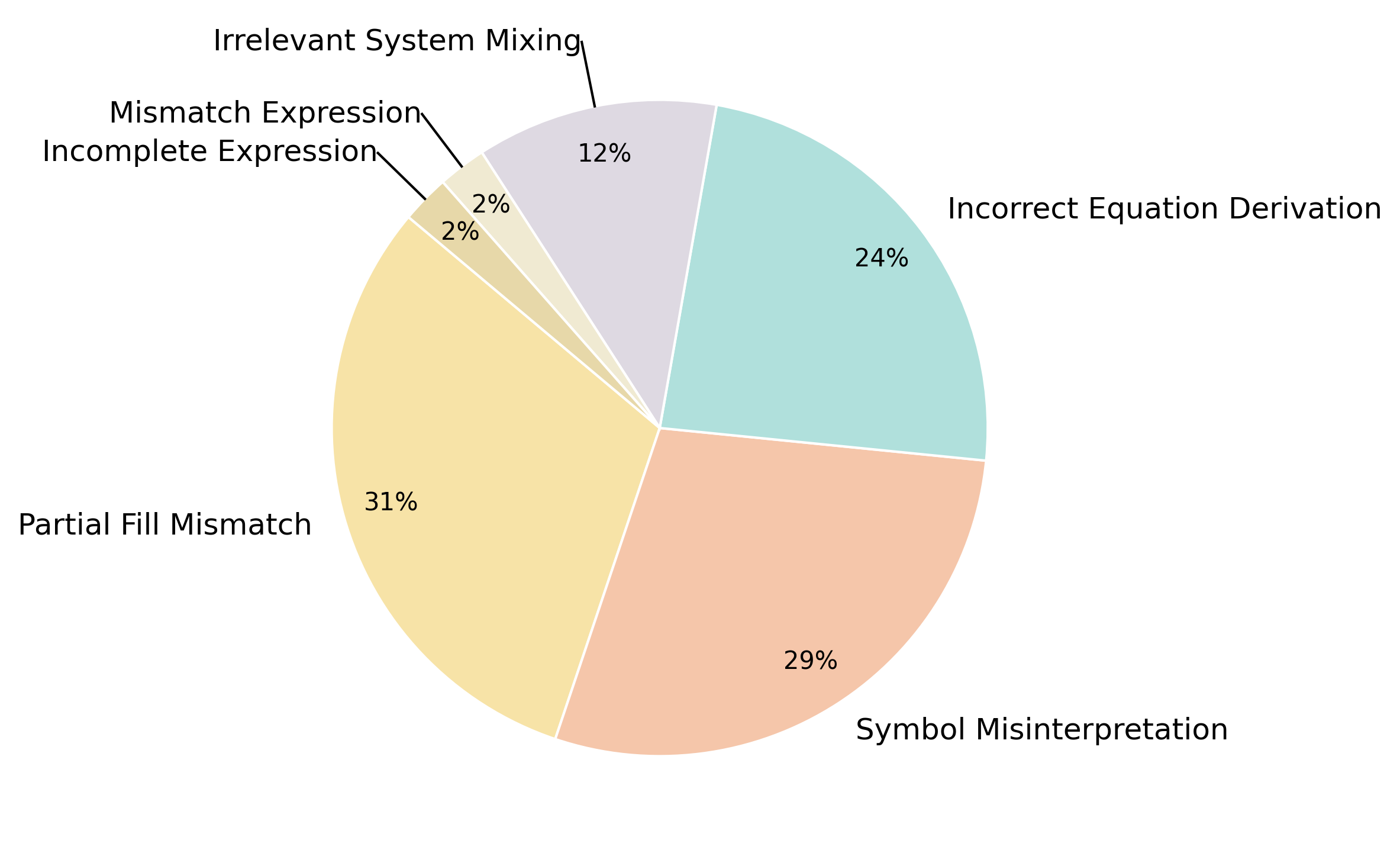}
    \caption{Error distribution among 40 annotated DeepSeek-R1 errors.}
    \label{fig:error_distribution}
\end{figure}

To better understand the limitations of the model-generated solutions, we randomly sampled and reviewed 40 failure answers by DeepSeek-R1 on the progressively masked filling and fully masked equation tasks, grouping them into several recurring categories. Figure~\ref{fig:error_distribution} summarizes the distribution of these errors.

\paragraph{Partial Fill Mismatch (31\%).}
A prevalent error pattern in progressive masking tasks involves models correctly completing one masked element while failing on others within the same expression. This manifests as either merging separate placeholders (e.g., combining $\sqrt{pK}\beta_{mk}$ and $\mathbf{y}_{pm}^H\pmb{\phi}_k$ into one term), misplacing correct expressions, or deriving one placeholder correctly while producing inconsistent expressions for interdependent variables. These errors persist despite explicit boundary indicators, suggesting fundamental limitations in LLMs' ability to maintain symbolic consistency across multiple related elements in complex wireless communication equations.

\paragraph{Symbol Misinterpretation (29\%).}
This type of error arises when the model chooses the wrong symbol or omits key symbolic elements in the final equation. 
An example is substituting \(\mathbf{H}_{\text{BR}}\) instead of \(\mathbf{H}_{\text{BR}}^H\) in a channel derivation.

\paragraph{Incorrect Equation Derivation (24\%).}
Several solutions fail to follow the correct derivation path, either missing crucial intermediate steps or injecting extraneous components. 
In longer sequences, a single early mistake (e.g., confusing pilot power \(p\) with user transmit power \(\rho_k\)) tends to propagate, causing the final expression to be structurally flawed despite appearing superficially similar.

\paragraph{Irrelevant System Mixing (11\%).}
We also observe instances in which the model introduces extraneous terms or assumes mismatched system settings. 
For example, it might inject NOMA-based interference factors into an RIS–MIMO scenario with no multi-user superposition, or switch to an entirely different beamforming constraint unconnected to the original problem statement.

\paragraph{Other Errors (4\%).}
A small fraction of errors are too context-specific to classify under the main categories. These include incomplete expressions---where the answer stops abruptly without filling the entire formula---and mismatched expressions that repeat known placeholders without substituting the correct variables.

Overall, while the majority of failures fall into coherent error patterns, it is evident that the model struggles when tasks require nuanced symbol-to-mask matching or integrative domain reasoning. 
Therefore, strengthening the model's ability to accurately derive reasoning and align domain knowledge is a key challenge for future improvements.

\section{Conclusion}
\label{sec:conclusion}

We introduced \textbf{\method}, the first benchmark that comprehensively evaluates LLMs’ abilities to domain-specific mathematical engineering tasks in wireless communications.
By presenting a broad range of tasks---from multiple-choice questions to progressively masked fill-in-the-blank and full equation completions---\method\ assesses how well models handle real-world wireless communications mathematical engineering challenges.
Our experiments show that, although many leading LLMs perform well on simpler question types, their ability to reconstruct equations deteriorates significantly when partial or full derivation is required, exposing a critical shortfall in current LLM-assisted scientific innovation.
Future work will expand the scope and complexity of these wireless challenges, with the aim of enhancing LLMs' mathematical reasoning and domain adaptation. By advancing their integration into the next-generation wireless systems, we ultimately strive toward the development of more capable, general-purpose AI solutions for scientific and engineering applications.

\section*{Ethical Considerations}
\label{sec:ethics}

This paper focuses on the development of a benchmark for evaluating language models on mathematical modeling tasks in wireless communications. 
The source data of \method\ is curated from open-access research papers, ensuring that the benchmark is built on publicly available information. Meanwhile, we resummarize the papers and anonymize the content to prevent any potential privacy concerns.
In experiments, we follow all licensing agreements and terms of service for the models evaluated, ensuring that our work is conducted in compliance with ethical guidelines.

\section*{Limitations}
While \method\ provides a comprehensive evaluation of LLMs on wireless mathematical modeling tasks, several limitations remain. 
First, it mainly covers text-based problems (e.g., symbolic derivations), missing other key data types like antenna diagrams, simulation plots, and Radio frequency (RF) measurements measurements, which are crucial for real-world wireless tasks.
Second, while \method\ spans topics from MIMO to RIS, it may not cover all emerging areas, such as quantum communication or terahertz systems.
Third, our automated evaluation checks the final symbolic equivalence and dimensionality plausibility but may miss incorrect reasoning at intermediate steps.
Lastly, all tests were done in a zero-shot setting. While this reflects real-world use, it does not explore whether fine-tuning or retrieval-based methods could improve results. Future versions of \method\ may include training splits to support domain adaptation and wireless-specific fine-tuning.

\subsubsection*{Acknowledgments}
This research is supported by the National Research Foundation, Singapore and Infocomm Media Development Authority under its Future Communications Research \& Development Programme (FCP-NTU-RG-2024-025). We thank the anonymous reviewers for their valuable feedback and constructive suggestions that have helped improve this paper. We also express our gratitude to the authors of the research papers from which we derived our benchmark examples, and to the domain experts who contributed to the validation and refinement of WirelessMathBench.

\bibliography{custom}

\appendix
\clearpage
\section{Dataset Details}
\label{sec:dataset_details}

\subsection{Topics and Papers Selection}
\label{app:topic_selection}

Our study addresses modern wireless communication challenges by selecting topics based on three key criteria that ensure both the academic rigor and practical relevance of our work. 
First, we target topics that have undergone peer review and have been accepted by prestigious journals such as IEEE Transactions on Wireless Communications (TWC), IEEE Transactions on Communications (TCOM), and IEEE Journal on Selected Areas in Communications (JSAC), as well as by top-tier conferences including IEEE International Conference on Communications(ICC) and IEEE Global Communications Conference (Globalcom), and for which corresponding arXiv versions are available. 
Second, we focus on communication system modeling that entails complex, multi-step mathematical derivations. These derivations are designed to closely mimic the challenges encountered in real-world wireless communication scenarios, capturing the intricate interplay between theoretical constructs and practical system constraints. 
Third, we ensure topic diversity by covering a wide range of wireless communication scenarios and problem domains. Specifically, our study encompasses seven major communication scenarios and six key problem areas, including interference management, spectrum optimization, network coding, and energy efficiency.

\subsection{ArXiv Data Processing}
Our data processing pipeline is similar with~\citep{maatouk2024tele}. First begins with the removal of all comments from the LaTeX files using Google's arXiv LaTeX Cleaner\footnote{\url{https://github.com/google-research/arxiv-latex-cleaner}}. 
We then parse the LaTeX source to extract the core technical content by separating the main text and mathematical expressions from non-essential elements such as comments, figures, and tables. For submissions comprising multiple files linked via \verb|\input| commands, we utilize the \texttt{latexpand} tool\footnote{\url{https://ctan.org/pkg/latexpand}} to flatten the document into a single file, ensuring all dependencies are resolved. To address the variability introduced by author-defined macros (e.g., via \verb|\newcommand| or \verb|\def|), we automatically expand these using the \texttt{de-macro}\footnote{\url{https://ctan.org/pkg/de-macro}}, replacing custom macros with their full definitions and normalizing all mathematical expressions to a consistent LaTeX format. 
Non-informative content such as acknowledgments and extensive bibliographies are removed to focus on technical material and to ensure anonymity in dataset construction by removing all author information from the articles. 

\section{Expert Validation Methodology}
\label{app:expert_validation}

This appendix details the expert validation process employed to ensure the quality and correctness of all questions in WirelessMathBench.

\subsection{Expert Qualifications}
\label{app:expert_qualifications}

Our expert review team comprised five individuals with established expertise in wireless communications:

\begin{itemize}
    \item \textbf{Senior Doctoral Student (1 member)}
    \begin{itemize}
        \item Approximately 5 years of research experience in wireless communications
        \item Multiple first-author publications in reputable journals (IEEE TWC, IEEE TVT)
    \end{itemize}
    
    \item \textbf{Postdoctoral Researchers (2 members)}
    \begin{itemize}
        \item Each with around 10 years of professional research experience focused on wireless communications
        \item Each has published over 10 first-author, peer-reviewed papers in top journals (IEEE TWC, IEEE JSAC, IEEE TVT)
        \item Both have served as reviewers for top-tier journals for many years
        \item Each has garnered more than 1,500 Google Scholar citations
    \end{itemize}
    
    \item \textbf{Senior Faculty Members (2 members)}
    \begin{itemize}
        \item Each with more than 25 years of research experience in wireless communications
        \item Extensive publication records with over 40,000 Google Scholar citations
        \item Leadership positions in research institutions and significant contributions to the field
    \end{itemize}
\end{itemize}

\subsection{Cross-Verification Protocol}
\label{app:cross_verification}

To ensure the highest quality of our benchmark, we implemented a rigorous cross-verification protocol:

\begin{enumerate}
    \item Each question in WirelessMathBench was reviewed by a minimum of two experts.
    \item Experts independently verified both the mathematical correctness and clarity of each question.
    \item When disagreements arose, the reviewers engaged in detailed discussions until consensus was reached.
    \item For particularly complex derivations, a third expert was consulted to provide additional verification.
    \item All multiple-choice distractors were examined to ensure they represented plausible but incorrect options.
\end{enumerate}

This protocol ensured that all questions accurately reflected real-world wireless communications challenges while maintaining clear formulation and unambiguous answers.

\subsection{Review Workflow}
\label{app:review_workflow}

The question development and verification process followed a structured workflow:

\begin{enumerate}
    \item \textbf{Initial Extraction:} Semi-automated extraction of system models from research papers.
    \item \textbf{Question Formulation:} Transformation of system models into question-answer pairs with varying difficulty levels.
    \item \textbf{First Review:} Initial expert review focusing on mathematical correctness, dimensional consistency, and domain applicability.
    \item \textbf{Refinement:} Modification of questions based on first review feedback.
    \item \textbf{Second Review:} Independent validation by a different expert, focusing on clarity and pedagogical value.
    \item \textbf{Consensus Discussion:} Resolution of any discrepancies between reviewer assessments.
    \item \textbf{Final Approval:} Acceptance of questions into the benchmark dataset after successfully passing all reviews.
\end{enumerate}

Throughout this process, reviewers paid particular attention to:
\begin{itemize}
    \item Dimensional consistency of all equations
    \item Proper use of notation and symbols
    \item Physical feasibility of the models
    \item Clarity and unambiguity of question formulation
    \item Appropriate difficulty level classification
\end{itemize}

This multi-stage review process ensured that WirelessMathBench contains high-quality questions that authentically represent the mathematical challenges in wireless communications.

\section{Prompt Templates}
\label{app:prompt_examples}

\begin{figure*}[!htbp]
    \centering
    \examplebox{Paper Summary Prompt Template}{
        \centering
        \includegraphics[width=0.6\linewidth]{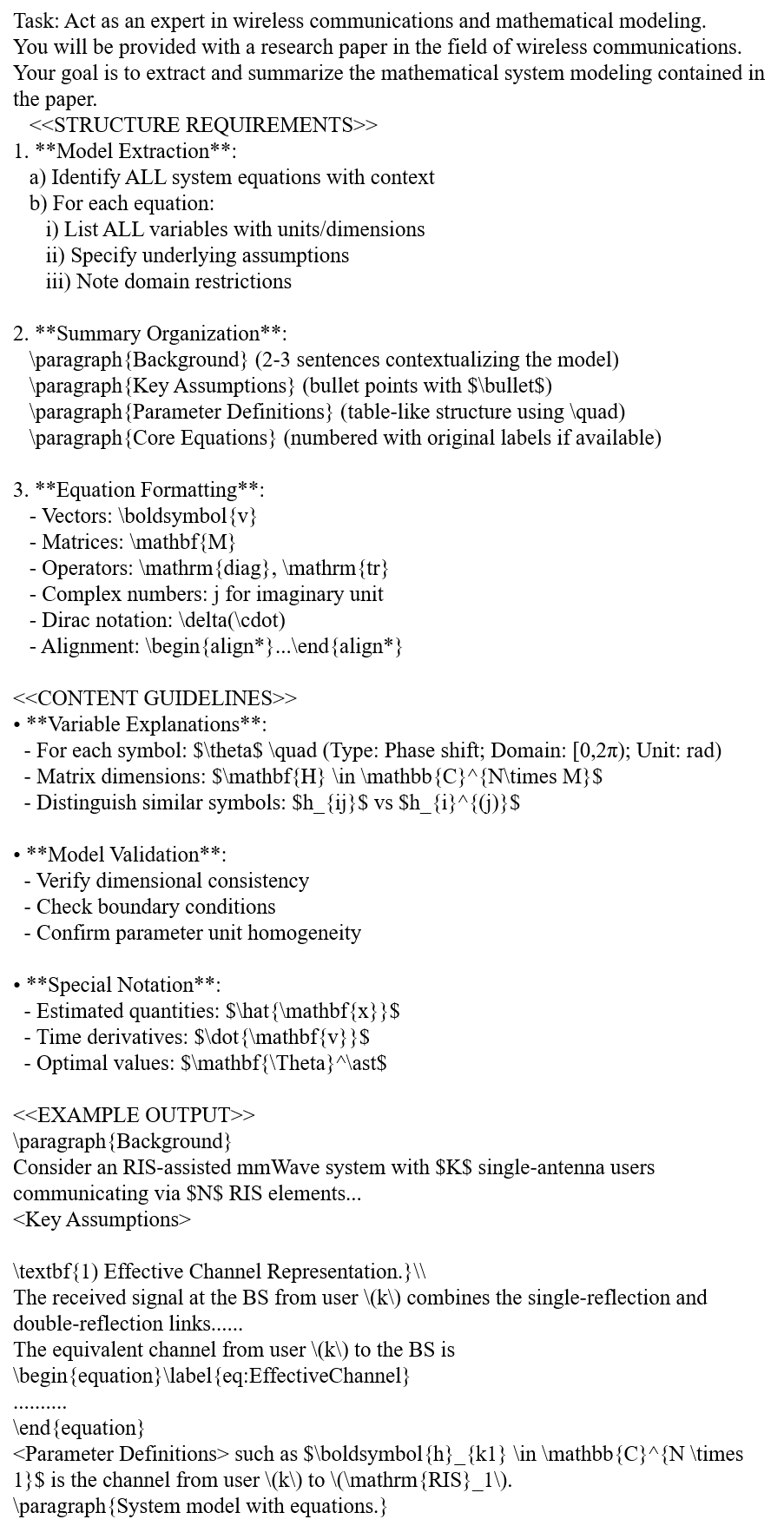}
    }
    \caption{This template is used to generate concise summaries of research papers.}
    \vspace{-1em}
    \label{fig:prompt:paper_summary}
\end{figure*}

\begin{figure*}[!htbp]
    \centering
    \examplebox{Question Generation Prompt Template}{
        \centering
        \includegraphics[width=0.5\linewidth]{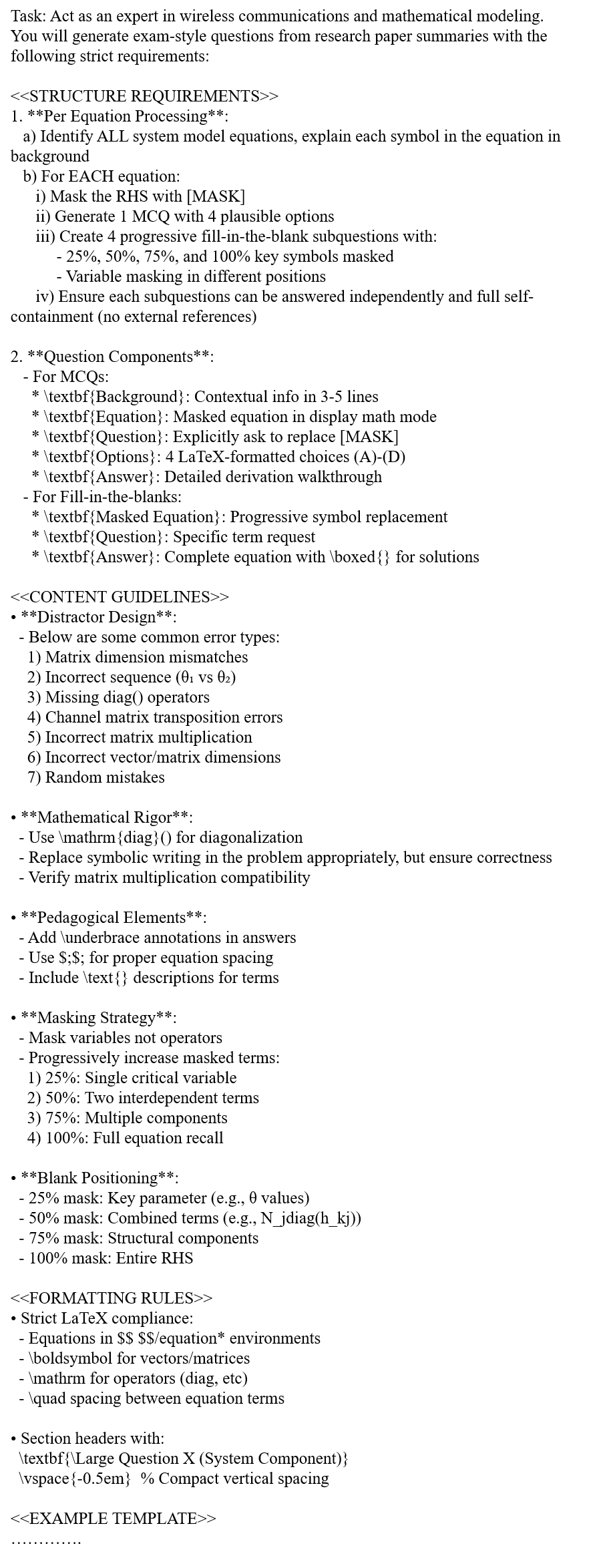}
    }
    \caption{This prompt template converts paper summaries into detailed question–answer pairs.}
    \vspace{-1em}
    \label{fig:prompt:question_generation}
\end{figure*}

\begin{figure*}[!htbp]
    \centering
    \examplebox{Multiple-Choice Question Prompt Template}{
        \centering
    \includegraphics[width=0.9\linewidth]{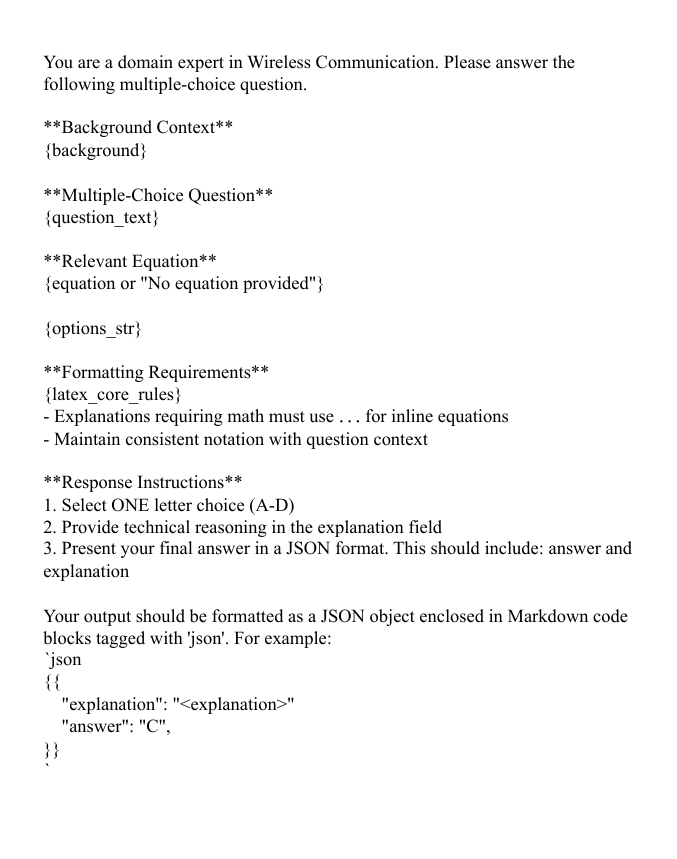}
    }
    \caption{This template is designed for answering multiple-choice questions. The model is guided to choose the correct mathematical expression from a set of closely related options.}
    \vspace{-1em}
    \label{fig:prompt:mcq}
\end{figure*}

\begin{figure*}[!htbp]
    \centering
    \examplebox{Fill-in-the-Blank and Full equation Completion Question Prompt Template}{
        \centering
    \includegraphics[width=0.9\linewidth]{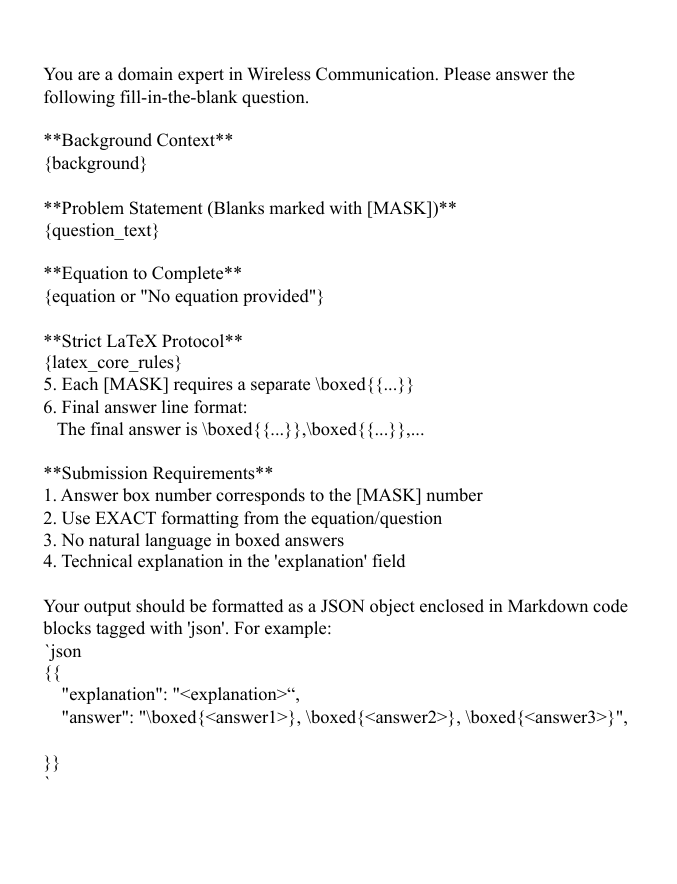}
    }
    \caption{Prompt template used for fill-in-the-blank and full equation completion tasks. It directs the model to reconstruct missing parts of equations using contextual cues and domain knowledge, simulating a step-by-step derivation process.}
    \vspace{-1em}
    \label{fig:prompt:fillin}
\end{figure*}

\begin{figure*}[!htbp]
    \centering
    \examplebox{LLM Evaluation Answer Prompt Template}{
        \centering
    \includegraphics[width=0.8\linewidth]{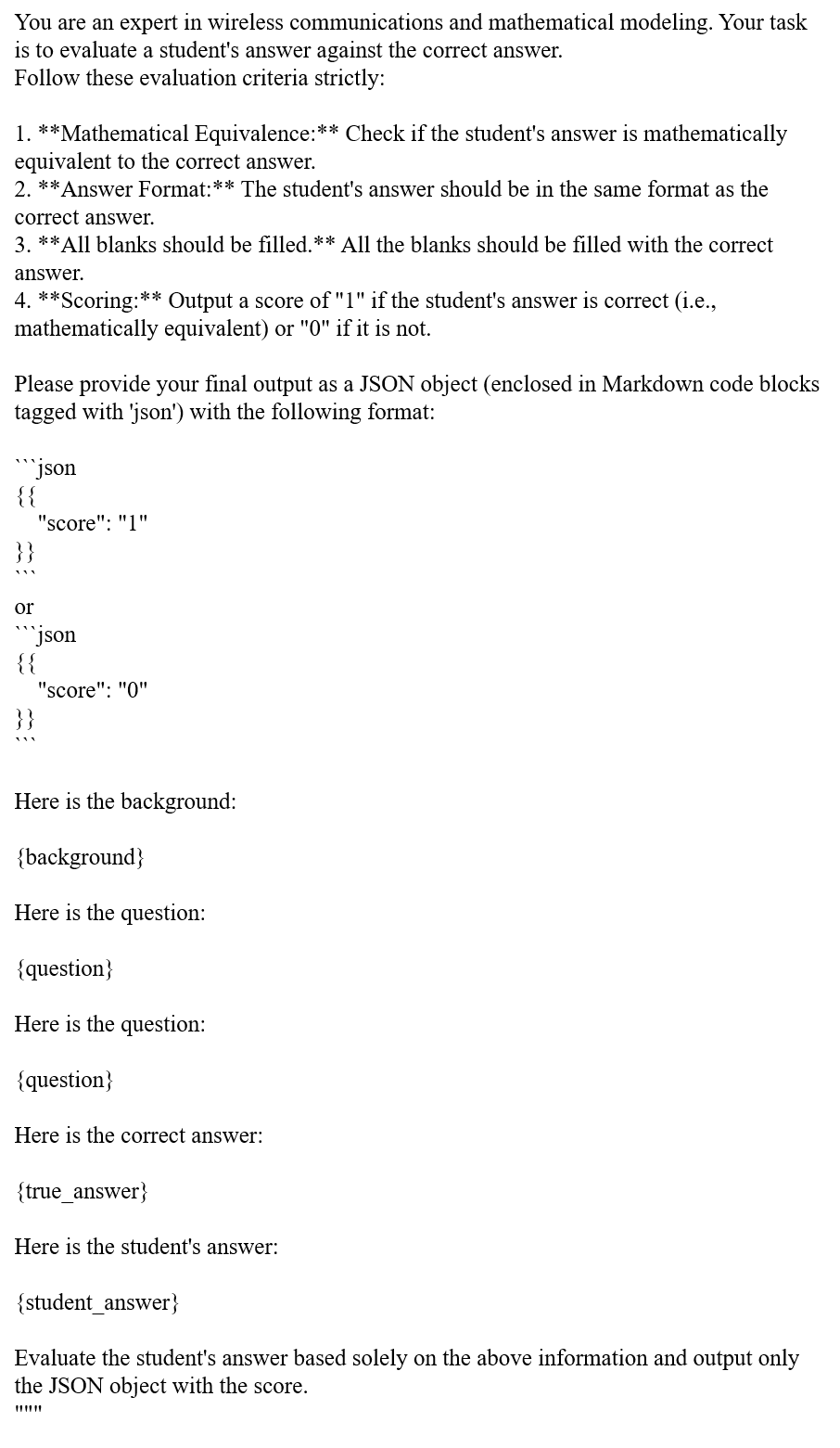}
    }
    \caption{This evaluation prompt template standardizes the process of assessing model-generated answers. It ensures that responses are judged consistently based on their correctness, completeness, and adherence to the required domain-specific reasoning.}
    \vspace{-1em}
    \label{fig:prompt:llm-judge}
\end{figure*}

For clarity and reproducibility, we provide examples of our prompt templates. Figure~\ref{fig:prompt:paper_summary} shows a template for the paper summarization task. Figure~\ref{fig:prompt:question_generation} illustrates the prompt used for question generation. Figures~\ref{fig:prompt:mcq} and~\ref{fig:prompt:fillin} present templates for multiple-choice and fill-in-the-blank questions, respectively. Finally, Figure~\ref{fig:prompt:llm-judge} displays the prompt used for LLM-based evaluation of model-generated answers.

\section{Model Configurations and Hyperparameters}

In our \method\ experiment, a total of 16 models were tested. All model tests followed the same template and set default parameters, and all results are the results of a single run.
 
\subsection{Closed-source Models}  
For models such as OpenAI-o1, GPT-4o, GPT-4, GPT-3.5-turbo, Gemini-2.0-flash, Gemini-1.5-pro, and Gemini-1.5-flash, we utilize their official API interfaces. These models are invoked via their respective API endpoints with standardized default parameters to ensure consistency and reproducibility across all experiments.

\subsection{Open-source Models}  
Our local experiments employ several open-source models deployed across different environments:
\begin{itemize}[leftmargin=*]
    \item \textbf{AliyunCloud Deployment\footnote{\url{https://bailian.console.aliyun.com/}}:} Qwen2.5-Math-72B and DeepSeek-V3 are deployed on the AliyunCloud platform.
    \item \textbf{NVIDIA NIM Deployment\footnote{\url{https://build.nvidia.com/}}:} DeepSeek-R1 and LLaMA-3.3-70B are run on NVIDIA NIM cloud platform.
    \item \textbf{HuggingFace Transformers\footnote{\url{https://huggingface.co/docs/transformers/en/index}}:} Other models---including Qwen2.5-Math-7B, LLaMA-3-8B-Tele, LLaMA-3-8B, and Mistral-7B---are run on local Nvidia A6000s, using the HuggingFace Transformers library to load pre-trained models.
\end{itemize}

\section{Example Output}
\label{app:example_outputs}

To demonstrate the gradation of complexity in \method{} and illustrate model performance across different task types, we present representative outputs from OpenAI-o1 and DeepSeek-R1, two high-performing models in our evaluation. 

\begin{figure*}[!htbp]
    \centering
    \examplebox{Q\_id:liu2023detecting\_q3\_mc}{
        \centering
        \includegraphics[width=\linewidth]{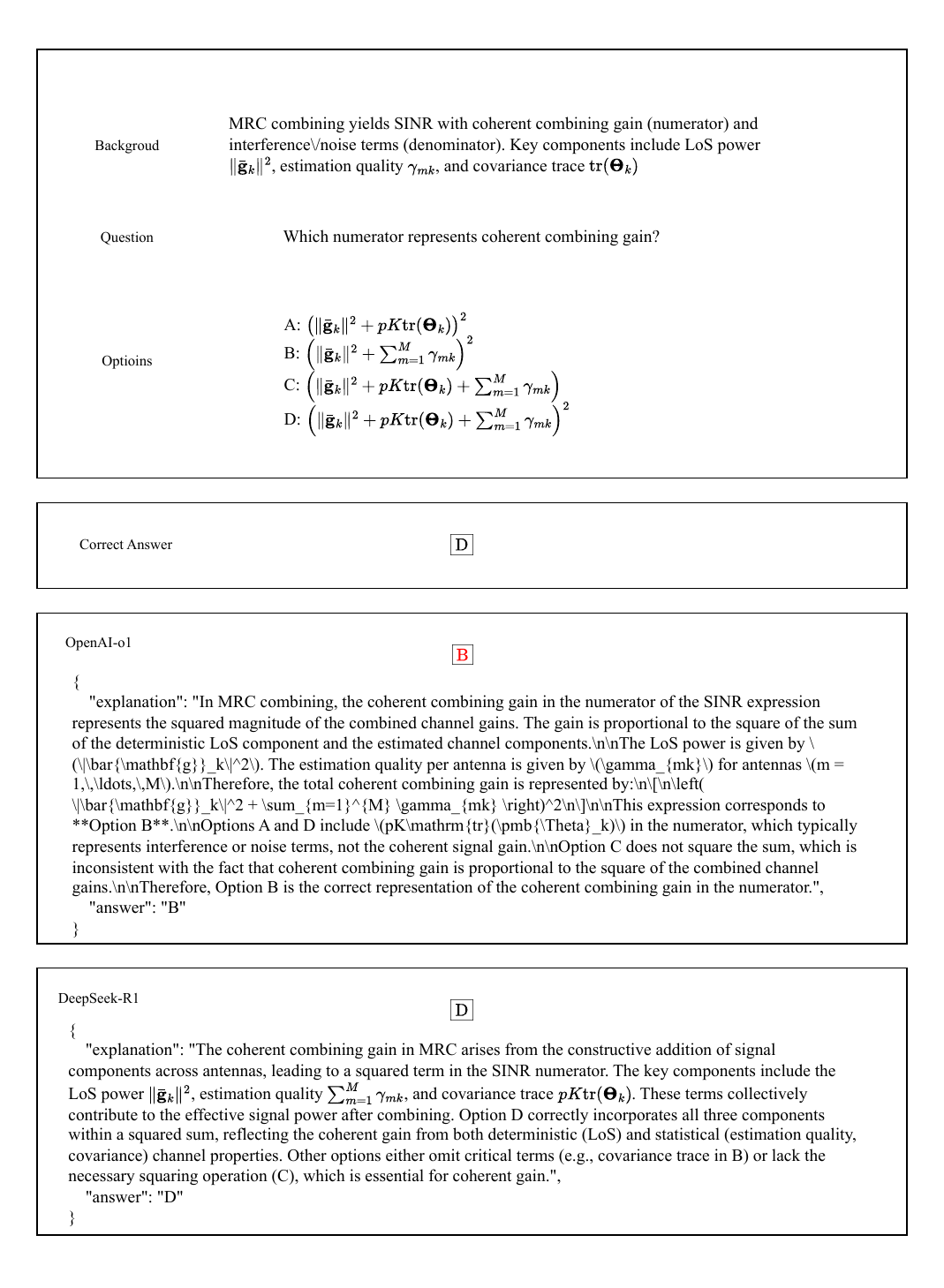}
    }
    \caption{An example question and the corresponding output from LLMs for a multiple-choice task.}
    \vspace{-1em}
    \label{fig:output_mc}
\end{figure*}

\begin{figure*}[!htbp]
    \centering
    \examplebox{Q\_id:liu2021deep\_q1\_fillin\_1}{
        \centering
        \includegraphics[width=\linewidth]{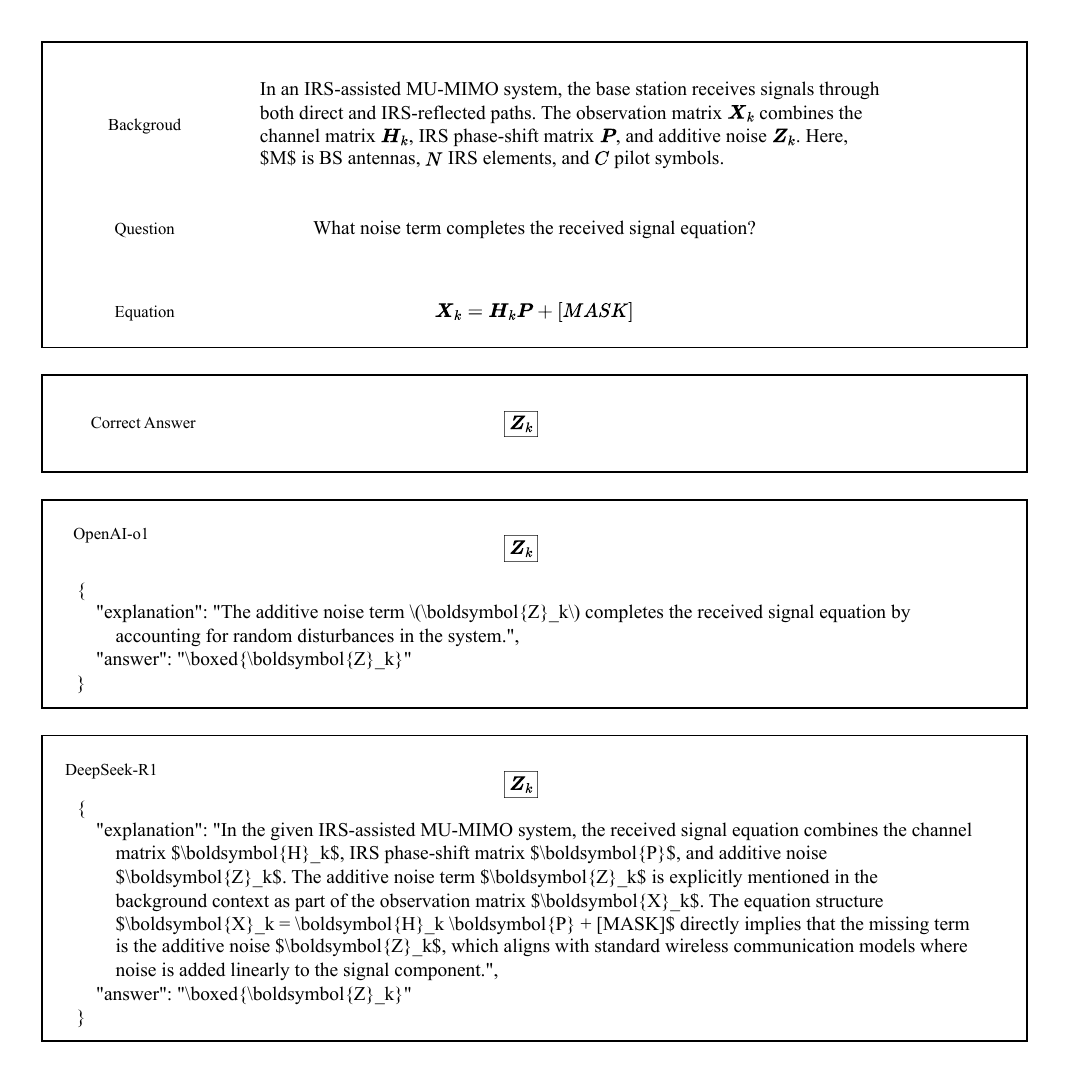}
    }
    \caption{An example of a level 1 progressively masked fill-in-the-blank task. The figure shows both the question and the LLM's output as it begins to reconstruct the missing parts of the equation.}
    \vspace{-1em}
    \label{fig:output_fillin1}
\end{figure*}

\begin{figure*}[!htbp]
    \centering
    \examplebox{Q\_id:zhao2024dual\_q1\_fillin\_2}{
        \centering
        \includegraphics[width=\linewidth]{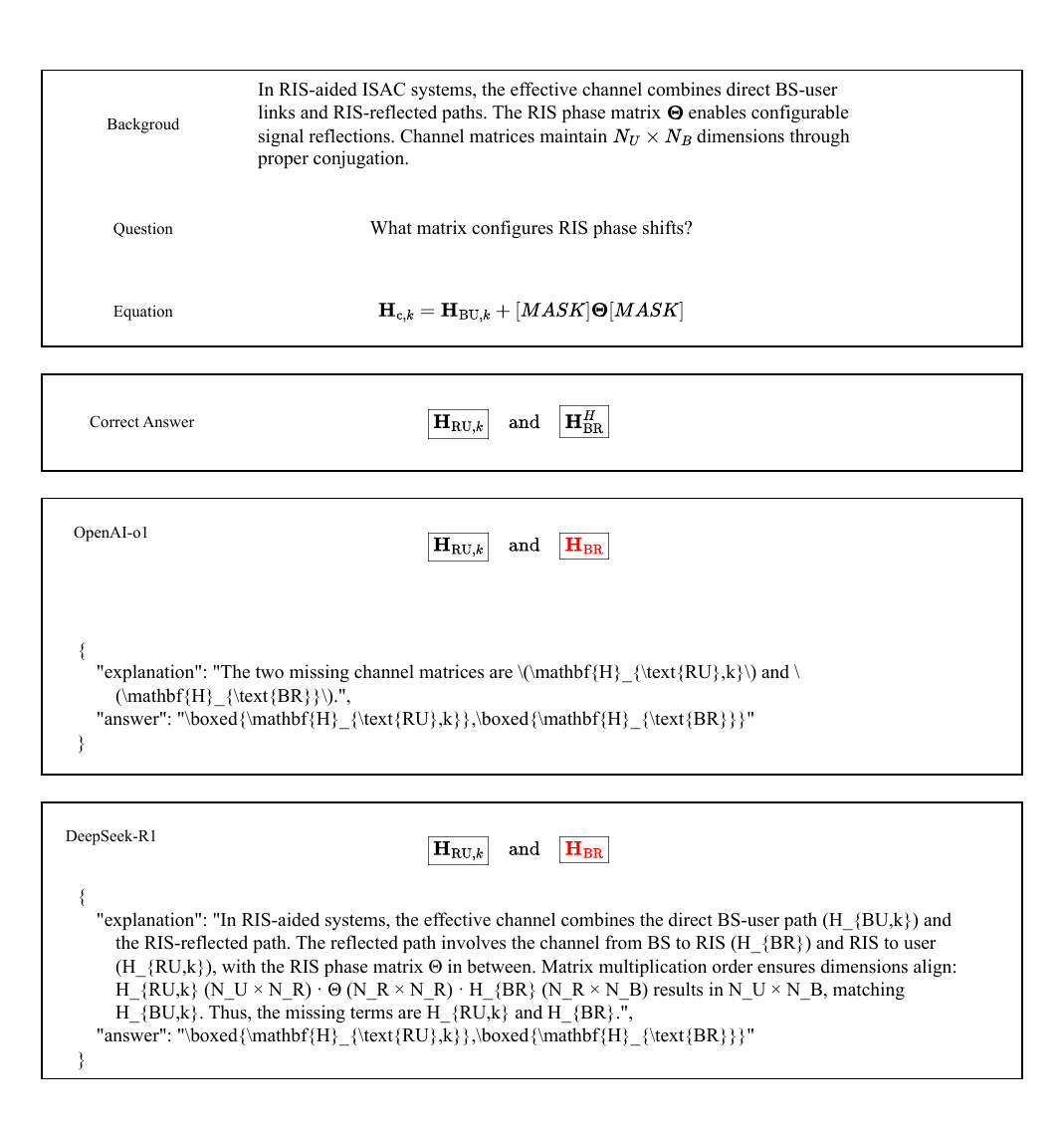}
    }
    \caption{An example of a level 2 progressively masked fill-in-the-blank task.}
    \vspace{-1em}
    \label{fig:output_fillin2}
\end{figure*}

\begin{figure*}[!htbp]
    \centering
    \examplebox{Q\_id:xu2021a\_q1\_fillin\_3}{
        \centering
        \includegraphics[width=\linewidth]{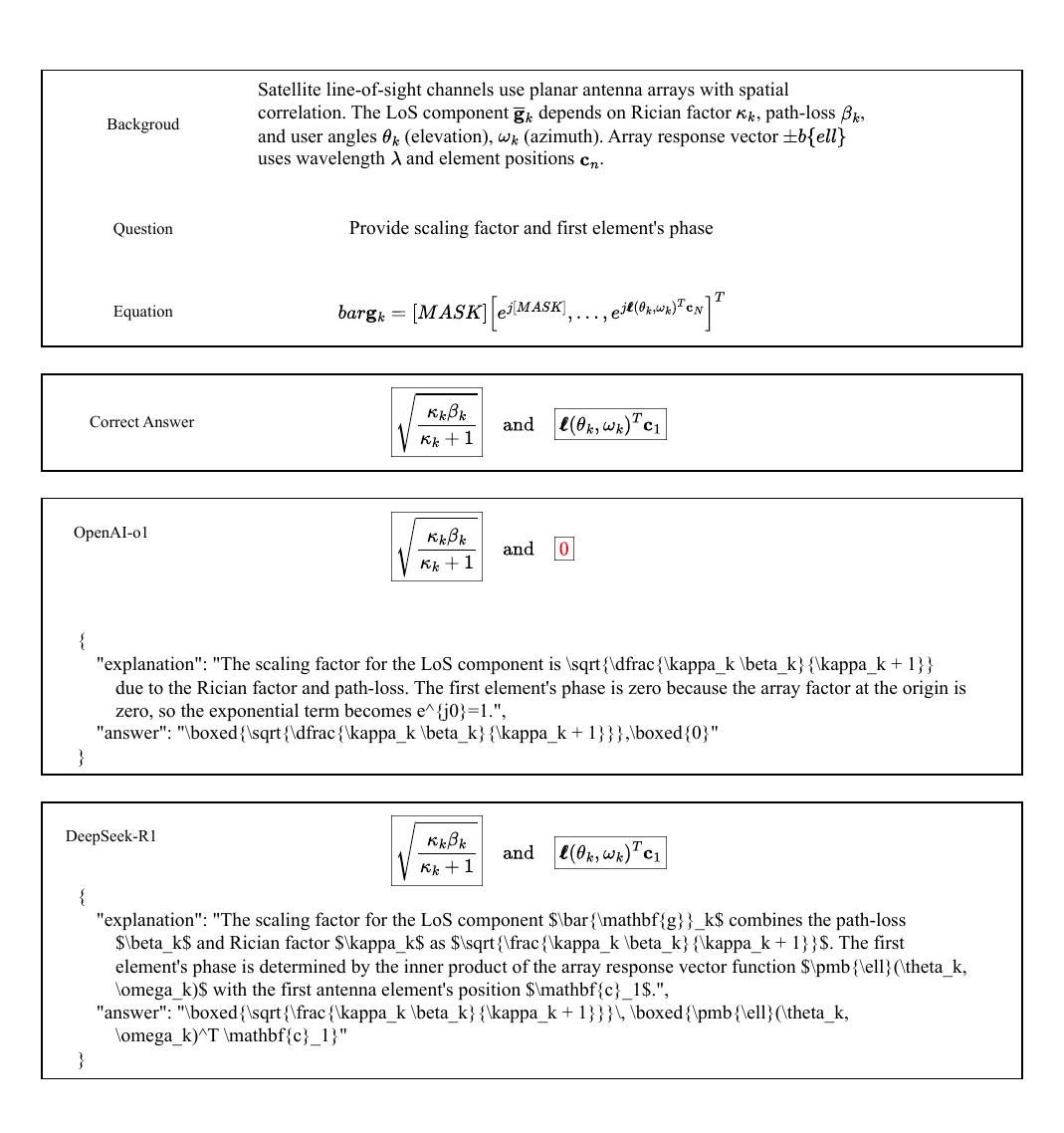}
    }
    \caption{An example of a level 3 progressively masked fill-in-the-blank task. Notice that both OpenAI-o1 and DeepSeek-R1 encountered difficulties in correctly interpreting the conjugate condition required by the task.}
    \vspace{-1em}
    \label{fig:output_fillin3}
\end{figure*}

\begin{figure*}[!htbp]
    \centering
    \examplebox{Q\_id:an2023stacked\_q1\_fillin\_4}{
        \centering
        \includegraphics[width=\linewidth]{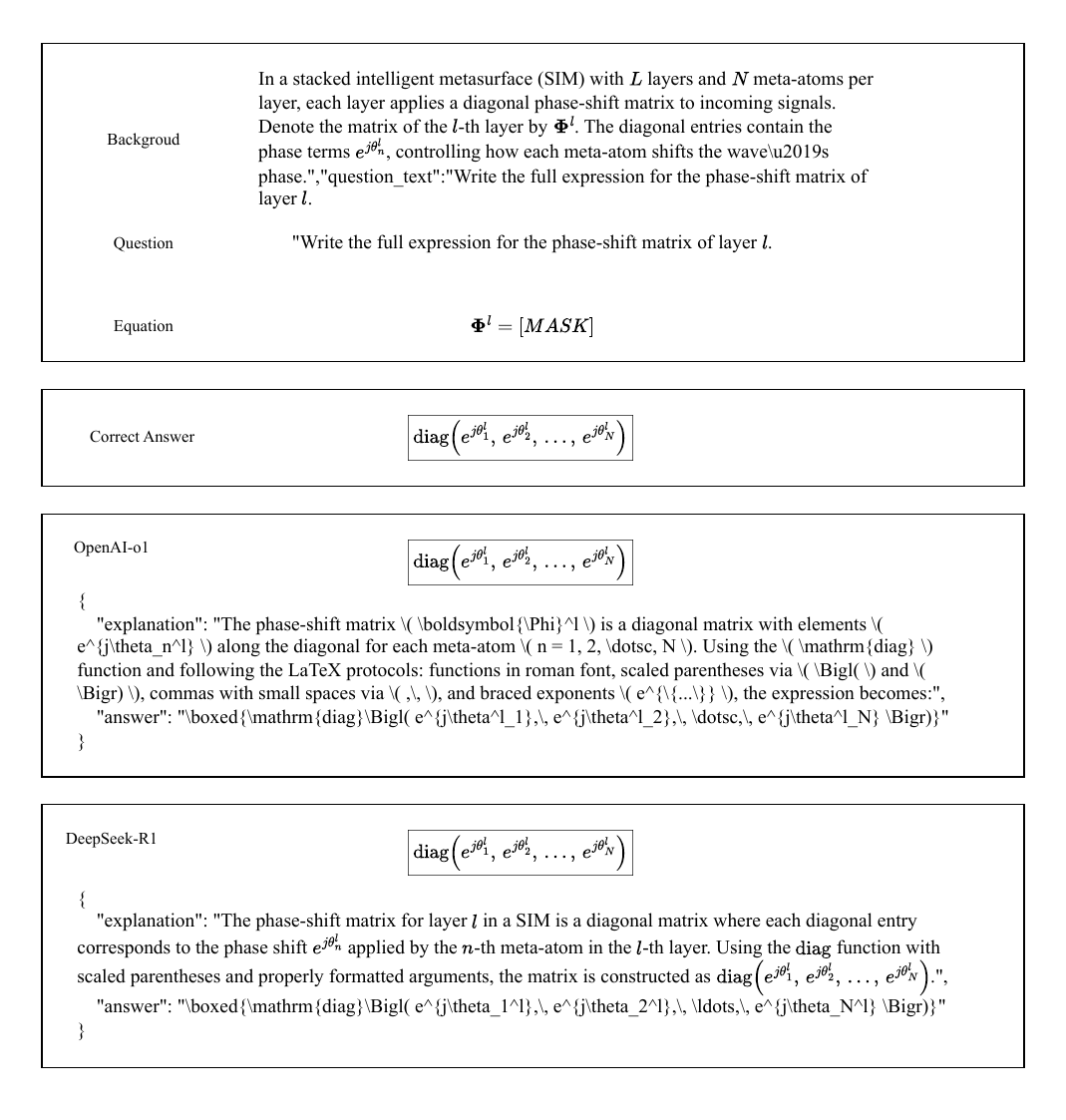}
    }
    \caption{An example of a Full Equation Completion (FEC) task.}
    \vspace{-1em}
    \label{fig:output_fec}
\end{figure*}

Figure~\ref{fig:output_mc} illustrates a multiple-choice task assessing MRC channel combining gain recognition, where models must select the correct mathematical formulation from similar distractors. Figures~\ref{fig:output_fillin1}, \ref{fig:output_fillin2}, and \ref{fig:output_fillin3} exemplify the progressive masking approach at increasing difficulty levels---from Level 1 (single variable substitution) to Level 3 (complex structured completion with conjugate conditions). Figure~\ref{fig:output_fec} demonstrates a Full Equation Completion task requiring the derivation of a phase-shift matrix formulation for stacked intelligent metasurfaces.

These examples corroborate our quantitative findings that model accuracy deteriorates substantially with increasing task complexity, even for state-of-the-art LLMs. Notably, all models struggle with complex dimensionality constraints and conjugation operations in higher-level tasks, suggesting fundamental limitations in their ability to maintain multi-step symbolic consistency in specialized engineering contexts.

\section{Chain-of-Thought Experiments}
\label{app:cot_experiments}
This section presents additional experiments exploring the impact of Chain-of-Thought (CoT) prompting on model performance in \method{}.

\subsection{Zero-shot Chain-of-Thought Prompt Template}
\label{app:cot_prompt}
For our CoT experiments, we modified our standard prompt templates to explicitly request step-by-step reasoning. The following shows the addition made to our base prompts:
\begin{quote}
    Provide a step-by-step explanation of your reasoning process before giving your final answer.
\end{quote}

\subsection{Performance Comparison}
\label{app:cot_performance}
Table~\ref{tab:cot_results} presents the comparison of model performance with and without Chain-of-Thought prompting across different task types.

\begin{table*}[!tb]
    \centering
    \begin{tabular}{lcccccc}
    \toprule
    \multirow{2}{*}{\textbf{Model}} & \multirow{2}{*}{\textbf{MCQ}}& \multicolumn{3}{c}{\textbf{Progressive Masking Filling}}&\multirow{2}{*}{\textbf{FEC}}&\multirow{2}{*}{\textbf{Avg. Acc}}\\
    
 \cmidrule(lr){3-5}&   & Level 1& Level 2& Level 3& & \\
    \midrule
 DeepSeek-R1& 76.00\%& \textbf{60.00\%}& \textbf{34.91\%}& \textbf{12.50\%}& \textbf{7.83\%}& \textbf{38.05\%}\\
 OpenAI-o1& 66.40\%& 59.17\%& 32.17\%& 8.04\%& 6.96\%&34.55\%\\
    \midrule
 GPT-4o& 72.80\%& 42.50\%& 28.70\%& 6.25\%& 4.35\%&30.92\%\\
 GPT-4o (w cot)& 72.00\% &	40.00\% &	23.48\% &	4.46\% &	5.22\% &	29.03\%\\
 GPT-4& 53.60\%& 38.33\%& 18.26\%& 3.57\%& 4.35\%&23.62\%\\
 GPT-4 (w cot)&	58.40\%&	32.50\%&	14.78\%&	3.57\%&	4.35\%&	22.72\% \\
 GPT-3.5-turbo& 45.60\%& 7.50\%& 10.43\%& 1.79\%& 1.74\%&13.41\%\\
 GPT-3.5-turbo (w cot)&	48.80\%&	11.67\%&	8.70\%&	2.68\%&	2.61\%&	14.89\% \\
 DeepSeek-V3& \textbf{78.40\%}& 50.00\%& 24.35\%& 6.25\%& 6.96\%&33.19\%\\
 DeepSeek-V3 (w cot)&	73.60\%&	50.00\%&	20.00\%&	6.25\%&	6.09\%&	31.19\% \\
    Gemini-2.0-flash& 71.20\%& 40.83\%& 24.35\%& 5.36\%& 4.35\%& 29.22\%\\
    Gemini-2.0-flash (w cot)&	73.60\%&	40.00\%&	24.35\%&	8.04\%&	4.35\%&	30.07\% \\
 Gemini-1.5-pro& 65.60\%& 43.33\%& 29.57\%& 9.82\%& 6.09\%&30.88\%\\
 Gemini-1.5-pro (w cot)&	69.60\%&	38.33\%&	25.22\%&	7.14\%&	3.48\%&	28.75\% \\
 Gemini-1.5-flash& 66.40\%& 37.50\%& 13.91\%& 2.68\%& 4.35\%&24.97\%\\
 Gemini-1.5-flash (w cot)&	68.80\%&	32.50\%	&17.39\%	&2.68\%&	5.22\%&	25.32\% \\
    \bottomrule
    \end{tabular}
    \caption{Performance comparison of models with and without Chain-of-Thought (CoT) prompting on WirelessMathBench.}
    \label{tab:cot_results}
\end{table*}

\subsection{Discussion of Chain-of-Thought Results}
\label{app:cot_discussion}
Our experiments with Chain-of-Thought (CoT) prompting yielded several interesting observations:
\begin{itemize}
    \item \textbf{Limited overall improvement:} Contrary to expectations, CoT prompting led to only marginal average accuracy improvements across most models, and in some cases, even decreased performance.
    
    \item \textbf{Task-dependent effects:} CoT showed small gains on simpler tasks (MCQ), but often hindered performance on more complex tasks (Level 2-3 masking and FEC). This suggests that explicit reasoning may introduce errors in complex symbolic manipulations when models lack robust mathematical reasoning capabilities.
    
    \item \textbf{Performance gap persistence:} Even with CoT prompting, models like GPT-4o and DeepSeek-V3 still significantly underperformed compared to models with built-in reasoning capabilities (DeepSeek-R1, OpenAI-o1), indicating that the ability to reason effectively about wireless communications problems cannot be induced solely through prompting.
    
    \item \textbf{Reasoning overflow:} In more complex tasks, we observed that CoT often led models to generate overly verbose reasoning chains that deviated from the critical path needed to solve the problem, potentially introducing errors.
\end{itemize}
These findings highlight that while CoT can offer some benefits for simpler tasks, addressing the fundamental challenges in complex mathematical modeling for wireless communications likely requires architectural improvements and specialized training rather than prompting strategies alone. Future work could explore more structured reasoning approaches or domain-specific fine-tuning to enhance performance on \method{}.

\section*{Disclosure}
In the process of writing this paper, we partially utilized ChatGPT as a language polishing tool to improve the clarity and quality of the text. However, all research ideas, data analyses, and conclusions were independently conceived and confirmed by the authors.

\newpage

\end{document}